%% file: main.tex
\documentclass{svproc}
\usepackage[T1]{fontenc}
\usepackage{graphicx}
\usepackage{booktabs}
\usepackage{amsmath}
\usepackage{wrapfig}
\usepackage{threeparttable}
\usepackage[numbers]{natbib}
\usepackage{float}
\usepackage{threeparttable}
\usepackage{placeins}
\usepackage{natbib}
\usepackage[colorlinks=true, linkcolor=blue, citecolor=blue, urlcolor=blue]{hyperref}

\usepackage{etoolbox}

\makeatletter
\patchcmd{\thebibliography}{\chapter*}{\section*}{}{}
\makeatother

\begin{document}
\title{Enhancing Regional Airbnb Trend Forecasting Using LLM-Based Embeddings of Accessibility and Human Mobility}
%
%
\author{Hongju Lee\inst{1} \and Youngjun Park\inst{2} \and Jisun An\inst{3} \and Dongman Lee\inst{2}}

\authorrunning{H. Lee et al.}
\titlerunning{Airbnb Trend Forecasting via LLM-Based Regional Embeddings}

\institute{
Graduate School of Data Science, KAIST, Daejeon, Republic of Korea, 34141 \\
\email{lhj5561@kaist.ac.kr}
\and
School of Computing, KAIST, Daejeon, Republic of Korea, 34141 \\
\email{\{youngjourpark, dlee\}@kaist.ac.kr}
\and
Luddy School of Informatics, Computing, and Engineering, Indiana University Bloomington (IUB), USA, 47405 \\
\email{jisunan@iu.edu}
}

\maketitle              

\input{section/abstract}
\input{section/introduction}
\input{section/relatedwork}
\input{section/model}

\input{section/experiment}

\input{section/conclusion}

\bibliographystyle{plain}
\bibliography{references}

%
%
%
%
\end{document}

%% file: section/abstract.tex
\begin{abstract}
The expansion of short-term rental platforms, such as Airbnb, has significantly disrupted local housing markets, often leading to increased rental prices and housing affordability issues. Accurately forecasting regional Airbnb market trends can thus offer critical insights for policymakers and urban planners aiming to mitigate these impacts. This study proposes a novel time-series forecasting framework to predict three key Airbnb indicators—Revenue, Reservation Days, and Number of Reservations—at the regional level. Using a sliding-window approach, the model forecasts trends 1 to 3 months ahead. Unlike prior studies that focus on individual listings at fixed time points, our approach constructs regional representations by integrating listing features with external contextual factors such as urban accessibility and human mobility. We convert structured tabular data into prompt-based inputs for a Large Language Model (LLM), producing comprehensive regional embeddings. These embeddings are then fed into advanced time-series models (RNN, LSTM, Transformer) to better capture complex spatio-temporal dynamics. Experiments on Seoul’s Airbnb dataset show that our method reduces both average RMSE and MAE by approximately 48\% compared to conventional baselines, including traditional statistical and machine learning models. Our framework not only improves forecasting accuracy but also offers practical insights for detecting oversupplied regions and supporting data-driven urban policy decisions.
\keywords{Airbnb, Regional Embedding, LLM Embedding, Time-series forecasting, Human mobility data}
\end{abstract}

%% file: section/Introduction.tex
\section{Introduction}


The recent rise of sharing economy platforms such as Airbnb has significantly reshaped local housing markets, triggering both academic research and policy debates on its socioeconomic implications.

Empirical studies have shown that regions with concentrated Airbnb activity experience significant increases in rental and housing prices. For instance, in Barcelona, every increase of 54 Airbnb listings corresponded to a 1.9\% rise in average rents and a 4.6\% rise in housing prices~\cite{garcia2020short}. Likewise, in Los Angeles, the enforcement of the Home Sharing Ordinance (HSO) led to a 50\% decrease in Airbnb listings and a subsequent 2\% reduction in both rent and housing values~\cite{koster2021short}. These findings collectively indicate that Airbnb's proliferation contributes to the displacement of long-term rentals in favor of short-term accommodations, intensifying pressures in the housing market. Thus, developing accurate forecasting methods for regional Airbnb trends has become crucial---not only for academic analysis but also for informing proactive urban policy interventions and regulatory actions aimed at stabilizing housing markets.

Previous machine learning studies have mainly focused on predicting Airbnb prices at the individual listing level, often using static features such as amenities, host characteristics, review counts, and proximity to points of interest (POI)~\cite{jiang2022multi, kirkos2022airbnb, sengupta2021examining, tan2024sustainable}. However, these approaches inherently neglect the spatio-temporal dynamics of the Airbnb market. Airbnb listings are typically concentrated near urban cores or tourist attractions, continuously influenced by dynamic factors such as human mobility patterns, transportation infrastructure, and urban accessibility. Moreover, housing market analyses typically require a broader regional context rather than isolated listings. Existing listing-level methods struggle to reflect these regional dynamics, making them insufficient for accurately predicting and responding to shifts in housing market conditions at a neighborhood scale.

To overcome these limitations, this study proposes a novel regional-level Airbnb trend forecasting framework based on multivariate time-series analysis. We utilize monthly Airbnb data aggregated at the smallest administrative unit (\textit{dong}) in Seoul, which allows for nuanced regional-level analysis aligned with previous housing market studies~\cite{franco2021impact, garcia2020short}. Our model targets three core indicators of Airbnb market activity: \textit{Revenue}, \textit{Reservation Days}, and \textit{Number of Reservations}. These indicators, widely recognized in prior research~\cite{kirkos2022airbnb, rubino2018airbnb, sengupta2021examining}, directly reflect Airbnb market performance and operational dynamics. 

Unlike traditional listing-focused approaches, our model explicitly integrates critical external regional context—specifically detailed transportation accessibility (including road networks, transit usage) and fine-grained human mobility patterns (such as domestic and international visitor flows)—to better capture dynamic spatial interactions influencing Airbnb market trends. To effectively encode these heterogeneous and complex contextual factors into a unified representation, we leverage the semantic and contextual understanding capabilities of Large Language Models (LLMs). Specifically, structured regional data are translated into natural-language prompts and processed through LLaMA 3~\cite{grattafiori2024llama}, producing rich embeddings that reflect intricate spatial and temporal dynamics which traditional numeric representations might fail to capture.

This study makes several notable contributions to Airbnb forecasting literature:

\begin{itemize}
    \item \textbf{(1) Regional-level Multivariate Forecasting}: Unlike prior studies primarily focusing on predictions at the individual listing level, we explicitly address the regional spatio-temporal dynamics of Airbnb markets using a multivariate time-series framework. Our method effectively captures spatial clustering and temporal dependencies inherent in urban and tourist-centric neighborhoods.

    \item \textbf{(2) Context-Aware Embeddings via LLM}: We propose a novel embedding strategy that leverages external regional context—specifically transportation accessibility and human mobility data—to generate semantically enriched embeddings through LLM-based prompting. This innovative approach substantially improves the representational capability of regional features compared to conventional numeric encoding methods.

    \item \textbf{(3) Robust and Generalizable Performance Improvements}: Empirical evaluations conducted on Seoul's Airbnb dataset demonstrate that our method consistently achieves an approximately 48\% reduction in both RMSE and MAE compared to conventional baseline models, including traditional statistical methods and standard machine learning approaches. These results validate the effectiveness and robustness of our context-aware embedding approach in capturing complex Airbnb market dynamics.
\end{itemize}

Overall, the proposed model offers not only methodological advancements in regional Airbnb trend prediction but also practical implications for identifying oversaturated areas, guiding regulatory interventions, and supporting evidence-based urban policy decisions.

%% file: section/relatedwork.tex
\section{Related Work}

\subsection{Socioeconomic Impacts of Airbnb on Local Housing Markets}

The rapid expansion of short-term rental platforms like Airbnb has had significant socioeconomic impacts on local housing markets, notably affecting rental prices and housing affordability. Multiple empirical studies have systematically explored these phenomena through spatially explicit regional analyses.

Garcia-López et al.~\cite{garcia2020short} utilized spatial econometric methods at Barcelona's Basic Statistical Area (BSA) scale, demonstrating that Airbnb listing density strongly correlated with increased rents and housing prices, particularly in tourism-centric regions. Similarly, Koster et al.~\cite{koster2021short} employed a Spatial Regression Discontinuity Design (Spatial RDD) to analyze the policy-induced reduction in Airbnb listings within Los Angeles, revealing a direct causal effect on local housing costs. Franco et al.~\cite{franco2021impact} also reported substantial price escalations linked to Airbnb density at the parish level in Portugal, further highlighting the spatial clustering effects of Airbnb activity on local housing markets.

Collectively, these studies consistently reveal that regional Airbnb density is directly associated with increased rental prices and substantial shifts in local housing market dynamics. Notably, they all employ spatially disaggregated, region-level analyses—an approach that closely aligns with the analytical perspective and research motivation of our study.

\subsection{Machine Learning Approaches for Airbnb Price Prediction}

Prior research on Airbnb price prediction has primarily employed machine learning techniques to estimate nightly listing rates using static, listing-level features.

Tan et al.~\cite{tan2024sustainable} integrated multimodal inputs, combining structured metadata, textual descriptions, and image data via deep learning architectures such as BERT and MobileNet, enhancing predictive accuracy. Lektorov et al.~\cite{lektorov2023airbnb} utilized user-generated review sentiment alongside host and listing attributes, demonstrating the predictive superiority of Support Vector Regression (SVR) among conventional machine learning methods. Jiang et al.~\cite{jiang2022multi} proposed a Multi-Source Information Embedding (MSIE) framework, incorporating structured data, textual reviews, and spatial points-of-interest (POI) information, significantly outperforming traditional ensemble methods like XGBoost and Random Forest.

Despite their methodological advancements, these studies predominantly focus on static, individual listing-level predictions and thus neglect spatio-temporal dynamics and broader regional contextual factors. Moreover, the explicit incorporation of human mobility and regional accessibility features remains largely unexplored. Our research directly addresses these limitations by proposing a regionally-scaled, time-series forecasting approach that integrates external contextual factors (e.g., accessibility and human mobility) through novel, semantically enriched LLM-based embeddings. This provides a more nuanced understanding of regional Airbnb trends than previous listing-centric models.

\subsection{Time-Series Forecasting Methods for Housing Market Analysis}

Various forecasting methods have been extensively applied to time-dependent phenomena similar to Airbnb market dynamics. Initial studies relied on statistical methods like ARIMA and SARIMA~\cite{box1970arima}. Subsequently, machine learning algorithms such as XGBoost~\cite{chen2016xgboost} and LightGBM~\cite{ke2017lightgbm} have been adopted to improve forecasting accuracy. Recently, deep learning methods, particularly Recurrent Neural Networks (RNN)~\cite{elman1990rnn}, Long Short-Term Memory (LSTM) networks~\cite{hochreiter1997lstm}, and Transformers~\cite{vaswani2017attention}, have gained prominence due to their capability to model complex temporal dependencies and non-linear patterns effectively.

However, existing forecasting models typically rely solely on structured numeric inputs and rarely integrate rich textual or contextual embeddings. Unlike these conventional methods, our approach leverages embeddings generated by LLMs, encoding regional contextual information derived from textual prompts. By integrating these contextually enriched embeddings with advanced deep learning architectures (RNN, LSTM, Transformer), our model achieves superior predictive accuracy and interpretability, better capturing both temporal and spatial nuances of Airbnb market dynamics.

\subsection{Research Gap and Our Contribution}

In summary, existing literature has largely overlooked the joint consideration of temporal dynamics, spatial context, and human mobility patterns in forecasting regional Airbnb market trends. Furthermore, no prior studies have effectively combined advanced time-series forecasting models with LLM-generated embeddings to capture nuanced regional contexts.

Our work bridges this gap by introducing a novel forecasting framework that integrates external contextual data through prompt-based LLM embeddings and advanced time-series models. This approach not only significantly improves prediction accuracy but also provides actionable insights for policy-making and urban management decisions related to housing markets.

%% file: section/model.tex
\section{Model}

\begin{figure}[htbp]
  \centering
  \includegraphics[width=\textwidth, height=0.45\textheight, keepaspectratio]{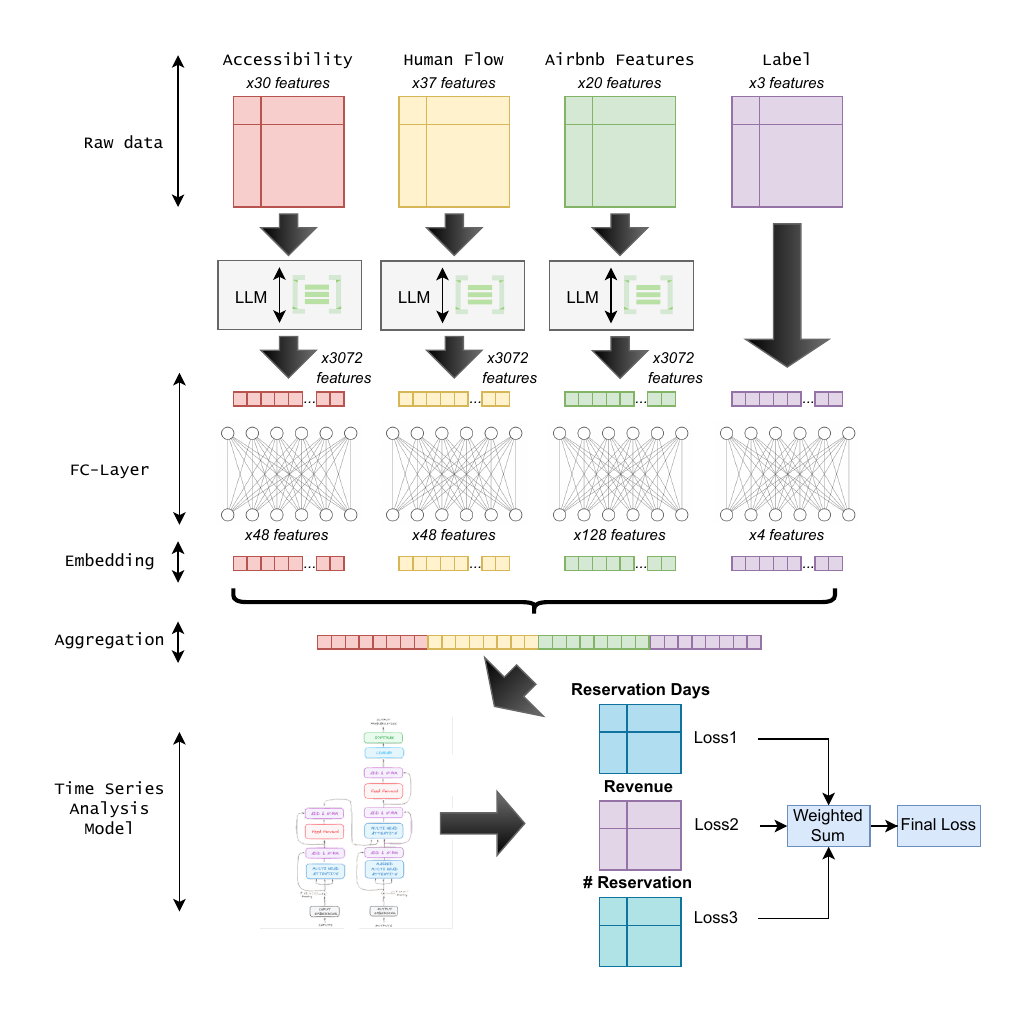}
  \caption{Overall architecture of the proposed Airbnb trend prediction model.}
  \label{fig:model}
\end{figure}

\subsection{Model Overview}

The overall architecture of the proposed model is depicted in Figure~\ref{fig:model}. The model leverages four distinct types of input data: Airbnb-specific features and label history, which have been traditionally used in prior studies ~\cite{kirkos2022airbnb, rubino2018airbnb, sengupta2021examining}, along with newly incorporated regional accessibility and human mobility data. These additional contextual features significantly enhance the model's capability to capture regional spatio-temporal variations. Detailed explanations of the input data are provided in Section~\ref{sec:model_input}. All input data are structured monthly at the \textit{dong} (administrative district) level, with label history specifically representing monthly values of three predictive targets: \textit{Reservation Days}, \textit{Revenue}, and \textit{Number of Reservations}.

Each input type is first converted into a semantically enriched embedding using an LLM, resulting in high-dimensional latent representations. These embeddings are subsequently reduced through fully connected (FC) layers into optimized lower-dimensional representations. The reduced embeddings are then concatenated to form comprehensive dong-level embeddings. Detailed procedures of this embedding process are described in Section~\ref{sec:llm_agg}.

The generated dong-level embeddings are organized into sequential inputs via a sliding-window mechanism, forming inputs for our multivariate time-series models. Each input sequence covers a historical period of six months. The model subsequently predicts three target indicators—\textit{Reservation Days}, \textit{Revenue}, and \textit{Number of Reservations}—for each dong over the subsequent three months following the window. During training, individual losses for each of the three prediction targets (Loss1, Loss2, Loss3) are computed and aggregated using a weighted sum to form the final loss function, which the model aims to minimize. Refer to Section~\ref{sec:window} for further details on the training and prediction procedures.

By effectively integrating external contextual information through LLM-generated embeddings with advanced time-series forecasting models, our proposed framework aims to accurately predict regional Airbnb market dynamics.

\subsection{Model Input Dataset} \label{sec:model_input}

To forecast regional Airbnb trends, we utilize three types of input data: Accessibility, Human Flow, and Airbnb Listings. These data sources reflect both the structural and social characteristics of each region and serve as the basis for constructing prompt-based LLM inputs. All data are aggregated monthly at the \textit{dong} (administrative district) level. The label history represents monthly values for three targets: \textit{Reservation Days}, \textit{Revenue}, and \textit{Number of Reservations}.

\begin{table}[ht]
\centering
\caption{Summary of Variables Used for Accessibility and Human Flow Prompts}
\label{tab:prompt_variables}
\resizebox{\textwidth}{!}{%
\begin{tabular}{ll}
\toprule
\textbf{Category} & \textbf{Variables} \\
\midrule
\textbf{Accessibility} & Road nodes near AirBnBs, Total roads and length, Tunnels, Bridges \\
& Counts of each road type, Bus and Subway ridership (on/off) \\
\midrule
\textbf{Human Flow} & Total domestic floating population, Domestic population by age and gender, \\
& Long-term foreign residents, \\
& Short-term foreign visitors \\
\bottomrule
\end{tabular}%
}
\end{table}

\subsubsection{Accessibility Embedding}

The accessibility embedding represents the ease of access to accommodations at the \textit{dong} (administrative district) level. It is constructed using data from OpenStreetMap (OSM)\footnote{\scriptsize Available at \url{https://www.openstreetmap.org/} (accessed 20 Apr 2025)} by aggregating various indicators related to road networks and transportation infrastructure at the dong level. Key features include the number of roads, the density of intersections within a 100-meter radius of Airbnb listings, and boarding/alighting passenger counts at bus and subway stations. These features quantitatively capture the structural accessibility of each region. 


\subsubsection{Human Flow Embedding}

The human flow embedding quantifies the mobility characteristics of each \textit{dong} using Seoul's living population data\footnote{\scriptsize
Data sources:
1) Domestic Living Population Data: \url{https://data.seoul.go.kr/dataList/OA-14991/S/1/datasetView.do}\\ 2) Long-Term Foreign Residents Data: \url{https://data.seoul.go.kr/dataList/OA-14992/S/1/datasetView.do}\\ 3) Short-Term Foreign Visitors Data: \url{https://data.seoul.go.kr/dataList/OA-14993/S/1/datasetView.do} (all accessed on 20 Apr 2025).
}.
 For domestic residents, the data includes the number of people segmented by gender and age group. For foreign populations, the data aggregates the living population of long-term and short-term stayers. All values are aggregated as monthly averages at the dong level to align with the temporal resolution of the prediction task. These variables are strongly associated with tourism density and inflow demand and thus function as key external factors in predicting Airbnb demand. The list of variables used to construct the accessibility and human flow prompts is summarized in Table~\ref{tab:prompt_variables}.

\subsubsection{Airbnb Embedding}

The Airbnb embedding is generated by aggregating structured listing data at the \textit{dong} level and computing statistical representations. The variables are selected based on prior studies~\cite{kirkos2022airbnb, rubino2018airbnb, sengupta2021examining} that demonstrated their effectiveness in predicting Airbnb revenue and demand. These include features related to property characteristics, pricing policies, operational information, and host attributes. For the baseline model, the structured data is processed into tabular format, with categorical variables one-hot encoded and numerical variables retained as-is. The full list of features used to construct the Airbnb embedding is presented in Table~\ref{tab:airbnb_features}. 

\begin{table}[ht]
\centering
\caption{List of Input Features for Airbnb Characteristics}
\resizebox{\textwidth}{!}{%
\begin{tabular}{ll}
\toprule
\textbf{Category} & \textbf{Feature Names} \\
\midrule
Operational Status & Available Days, Blocked Days \\
Accommodation Characteristics & Property Type, Listing Type, Bedrooms, Bathrooms, Max Guests \\
Host Information & Response Rate, Airbnb Response Time, Airbnb Superhost \\
Price/Policy & Cancellation Policy\\
Check-in Information & Check-in Time, Checkout Time, Minimum Stay \\
Others & Number of Photos, Instantbook Enabled, Pets Allowed, Integrated Property Manager\\
User Response & Overall Rating, Number of Reviews \\
\bottomrule
\end{tabular}%
}
\label{tab:airbnb_features}
\end{table}

\subsection{LLM Embedding \& Aggregation} \label{sec:llm_agg}

Instead of traditional statistical encoding, we convert the three regional-level datasets (Accessibility, Human Flow, and Airbnb Characteristics) into LLM-based embeddings. Specifically, we reformulate each \textit{dong}-month data record into a text-based prompt, which is then encoded into a 3072-dimensional embedding using LLaMA 3.

\subsubsection{Prompt Construction}

The prompts are constructed differently depending on the characteristics of each data type, specifically distinguishing numerical, binary, and categorical variables:

\begin{itemize}
    \item \textbf{Accessibility and Human Flow Data}: These datasets primarily consist of numerical variables, thus each prompt was created by listing the variable names alongside their numerical values. For example, an Accessibility prompt includes sentences such as "Total number of roads in the dong: 102, Total length: 15032.77" while a Human Flow prompt includes sentences like "Domestic Floating Population by Age and Gender 20s Male: 687.54, Female: 555.01" 

    \item \textbf{Airbnb Characteristics Data}: Due to the mixed nature of the data, the variables are processed differently:
    \begin{itemize}
        \item \textit{Binary variables}: Summarized by counting the number of true values.
        \item \textit{Categorical variables}: Summarized by counting occurrences per category.
        \item \textit{Numerical variables}: Summarized using statistics such as mean, standard deviation, maximum, and minimum.
    \end{itemize}
    The Airbnb listing features are formatted into prompts following a structured template, as shown below:
    
    {\scriptsize
    \begin{verbatim}
    [{month} | {dong}] Airbnb Feature Summary:
    Total number of AirBnBs: {count}
    
    Category Column Attributes:
    Category: {category name} Information: Total number with data: {count}
      - {value 1}: {count}
      - {value 2}: {count}
      ...
    (repeated for each categorical column)
    -----------------------------------------------------------
    Binary Column Attributes:
    {Binary Column name} Information: Total number with data: {count}
      - Number of {Binary Column name}: {True count}
    (repeated for each binary column)
    -----------------------------------------------------------
    Numerical Column Attributes:
    {Numerical Column name} Information: Total number with data: {count}
      - Mean: {mean}
      - Std Dev: {std}
      - Median: {median}
      - Min: {min}
      - Max: {max}
    (repeated for each numerical column)
    \end{verbatim}
    }
\end{itemize}

The resulting prompts encapsulate rich regional contexts at each \textit{dong}-month level and are subsequently encoded into high-dimensional embeddings through LLaMA 3.

\subsubsection{Embedding Reduction and Aggregation}

The generated 3072-dimensional LLM embeddings undergo a dimensionality reduction process to serve as inputs for the time-series forecasting model. As illustrated in the FC-layer block of Figure~\ref{fig:model}, each embedding is passed through a 4-layer fully connected neural network (3072 $\rightarrow$ 768 $\rightarrow$ 256 $\rightarrow$ 128 $\rightarrow$ output). Based on empirical experimentation, the final embedding dimensions  are set as follows: Accessibility Embedding (48 dimensions), Human Flow Embedding (48 dimensions), and Airbnb Characteristics Embedding (128 dimensions).

Finally, these embeddings are concatenated at the \textit{dong} level to form an integrated regional embedding. In addition, historical label values (\textit{Revenue, Reservation Days, and Number of Reservations}) are included in the time-series input to capture temporal patterns more effectively. Although the original label vector has 3 dimensions, it is expanded to 4 dimensions through a 1-layer fully connected layer to maintain uniformity in the input sequence.

\subsection{Sliding Window-based Time Series Construction}\label{sec:window}

The dataset is organized on a monthly basis, and we utilize a sliding window approach to perform time-series forecasting. Due to the limited size of the dataset, a stride of 1 is used, shifting the window forward by one month at each time step to maximize the number of training samples.

The input to the model consists of sequential embeddings of each \textit{dong} across the defined window size. The input tensor for each training sample has a shape of $\left( \text{window size}, N, D \right)$, where $N$ is the number of administrative regions (\textit{dongs}) and $D$ is the dimension of the concatenated regional embedding. At each prediction step, the model outputs forecasts for the next 1, 2, and 3 months simultaneously for three target variables—\textit{Revenue}, \textit{Reservation Days}, and \textit{Number of Reservations}. To match this objective, the output tensor has a shape of $\left( 3, N, 3 \right)$, corresponding to 3 future months, $N$ regions, and 3 target metrics. To train the model, we minimize a weighted sum of individual loss functions for the three prediction targets:

\begin{equation}
L_{\text{total}} = \alpha \cdot L_{\text{reservation days}} + \beta \cdot L_{\text{revenue}} + \gamma \cdot L_{\text{number of reservations}}
\end{equation}

As shown in the final output block of Figure~\ref{fig:model}, this loss calculation integrates predictions for all three target variables over the next three months.  
The weights $\alpha, \beta, \gamma$ are determined empirically based on validation performance, and are set equally in our experiments unless otherwise specified.

%% file: section/experiment.tex
\section{Experiments}

\subsection{Data Collection and Preprocessing}

\begin{wrapfigure}{r}{0.35\textwidth}
  \vspace{-10pt}  
  \centering
  \includegraphics[width=\linewidth]{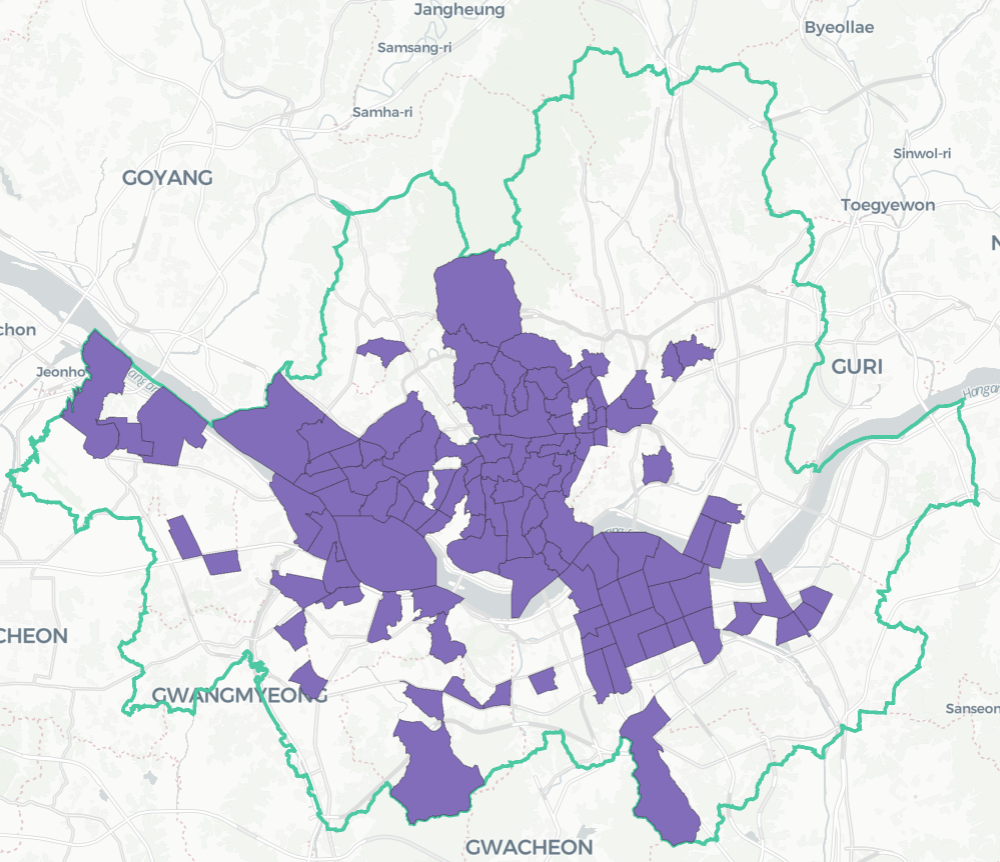}
  \caption{Top 25\% Most Active Airbnb Areas in Seoul}
  \label{fig:wrap_airbnb}
  \vspace{-10pt}
\end{wrapfigure}

We collect monthly Airbnb listing data from 424 administrative districts (\textit{dongs}) in Seoul, spanning from 2017 to 2022 (67 months in total). The data is sourced from AirDNA\footnote{\scriptsize Available at \url{https://www.airdna.co/} (accessed 20 Apr 2025)}, which provides monthly hosting information related to Airbnb listings~\citep{wang2023challenges}. To ensure data quality and temporal stability, we select only the top 25\% of the dongs (109) whose average number of listings exceeded the third quartile (22) throughout the period.

Figure~\ref{fig:wrap_airbnb} visualizes the selected regions, 
mainly in Seoul's major urban centers. These areas exhibit high tourist inflow and lodging density, reflecting the spatial clustering tendency of Airbnb listings.

We adopt a 6-month sliding window with a stride of 1 for time-series forecasting. The data set is chronologically divided into training (first 51 months), validation (next 8 months), and testing (final 8 months). Z-score normalization is applied based on the training set. To stabilize the variance in the target variables (\textit{Revenue}, \textit{Reservation Days}, and \textit{Number of Reservations}), we applied a logarithmic transformation to the labels prior to normalization. A fixed random seed of 43 was used to ensure reproducibility.

\subsection{Model Settings}

We use the LSTM model for hyperparameter search, as it shows the most stable performance with the lowest variance across repeated experiments.

\subsubsection{Embedding Dimension}

To determine the optimal embedding size for each input type--Accessibility, Human Flow, and Airbnb--we test four embedding configurations ($\text{D}_{\text{opt}1}$ to $\text{D}_{\text{opt}4}$), with the total dimensionality constrained to 244 due to computational limitations. As summarized in Table~\ref{tab:embedding_perf}, the configuration $\text{D}_{\text{opt}2}$ (48/48/128/4) achieves the best performance in terms of both RMSE and MAE. This result highlights the high information density of Airbnb features, as $\text{D}_{\text{opt}2}$ allocates more dimensions to the Airbnb embedding compared to $\text{D}_{\text{opt}1}$ (48/48/64/4), which performs noticeably worse. Additional variations in the auxiliary embeddings ($\text{D}_{\text{opt}3}$ and $\text{D}_{\text{opt}4}$) do not result in meaningful improvements. Based on these findings, we select $\text{D}_{\text{opt}2}$ as the final configuration for all subsequent experiments.

\begin{table}[htbp]
\centering
\begin{minipage}[t]{0.47\textwidth}
\centering
\begin{threeparttable}
\caption{Performance Comparison by Embedding Dimension Settings}
\begin{tabular}{lcc}
\toprule
\textbf{Embedding Option} & \textbf{RMSE} & \textbf{MAE} \\
\midrule
$\text{D}_{\text{opt}1}$ (48 / 48 / 64 / 4)   & 0.4630 & 0.3824 \\
$\text{D}_{\text{opt}2}$ (48 / 48 / 128 / 4)  & \textbf{0.4075} & \textbf{0.3243} \\
$\text{D}_{\text{opt}3}$ (48 / 64 / 128 / 4)  & 0.4287 & 0.3456 \\
$\text{D}_{\text{opt}4}$ (64 / 48 / 128 / 4)  & 0.4264 & 0.3429 \\
\bottomrule
\end{tabular}
\begin{tablenotes}
\small
\item \textit{Note.} Embedding configuration is defined as (Accessibility / Human Flow / Airbnb / Label).
\end{tablenotes}
\label{tab:embedding_perf}
\end{threeparttable}
\end{minipage}%
\hfill
\begin{minipage}[t]{0.47\textwidth}
\centering
\begin{threeparttable}
\caption{Prediction Performance by Window Size}
\begin{tabular}{lcc}
\toprule
\textbf{Window Size} & \textbf{RMSE} & \textbf{MAE} \\
\midrule
3 months  & 0.4229 & 0.3385 \\
6 months  & \textbf{0.4075} & \textbf{0.3243} \\
9 months  & 0.4115 & 0.3295 \\
12 months & 0.4334 & 0.3492 \\
\bottomrule
\end{tabular}
\label{tab:window_perf}
\end{threeparttable}
\end{minipage}
\end{table}

\subsubsection{Window Size Selection}

We evaluate the impact of input sequence length by comparing four different time window sizes: 3, 6, 9, and 12 months. As shown in Table~\ref{tab:window_perf}, the 6-month window consistently produces the lowest RMSE and MAE across experiments. While shorter windows lack sufficient temporal context and longer windows may introduce noise or outdated patterns, the 6-month window offers a balance between recency and trend stability. Consequently, we adopt the 6-month window size as the default setting for all model configurations.

\subsection{Main Results}

We evaluate whether a time-series forecasting model that integrates regional-level external information and LLM-based embeddings outperforms traditional approaches based solely on Airbnb listing features.
For comparison, we constructed a baseline model using both listing features—commonly employed in previous studies~\cite{kirkos2022airbnb, rubino2018airbnb, sengupta2021examining}—and label history data. The listing features are aggregated at the dong level, and categorical variables are one-hot encoded into a tabular format. To ensure a fair comparison, all models share the same architecture and output format, with embedding dimensions fixed to 128 for Airbnb features and 4 for label history.

\begin{table}[htbp]
\centering
\begin{threeparttable}
\caption{Prediction Performance across Models}
\label{tab:main_results}

\begin{tabular}{lcccccccc}
\toprule
\textbf{Model} & \multicolumn{2}{c}{\textbf{Total}} & \multicolumn{2}{c}{\textbf{Res. Days}} & \multicolumn{2}{c}{\textbf{Revenue}} & \multicolumn{2}{c}{\textbf{\#Reservations}} \\
               & RMSE & MAE & RMSE & MAE & RMSE & MAE & RMSE & MAE \\
\midrule
RNN Baseline         & 0.6635 & 0.4955 & 0.6575 & 0.4828 & 0.6723 & 0.5113 & 0.6608 & 0.4922 \\
LSTM Baseline        & 0.8090 & 0.6457 & 0.7809 & 0.6024 & 0.8091 & 0.6561 & 0.8371 & 0.6785 \\
Transformer Baseline & 0.9954 & 0.8606 & 0.9284 & 0.7779 & 1.1040 & 0.9840 & 0.9537 & 0.8198 \\
\midrule
RNN Our Model          & 0.4103 & 0.3271 & 0.3865 & 0.2992 & 0.3979 & 0.3222 & 0.4465 & 0.3599 \\
LSTM Our Model         & \textbf{0.4075} & \textbf{0.3243} & \textbf{0.4000} & \textbf{0.3168} & 0.3976 & 0.3189 & \textbf{0.4249} & \textbf{0.3374} \\
Transformer Our Model  & 0.4240 & 0.3439 & 0.4360 & 0.3569 & \textbf{0.3502} & \textbf{0.2756} & 0.4859 & 0.3991 \\
\bottomrule
\end{tabular}

\begin{tablenotes}
\small
\item \textit{Note.} Res. Days = Reservation Days; \#Reservations = Number of Reservations. \\ Baseline models are constructed using simple statistical aggregation of Airbnb listing features at the dong level.

\end{tablenotes}
\end{threeparttable}
\end{table}

As shown in Table~\ref{tab:main_results}, our proposed model consistently outperforms the baseline across all time-series architectures---RNN, LSTM, and Transformer---across all evaluation metrics. The LSTM-based version of our model achieves the best overall performance, with the lowest total RMSE (0.4075) and MAE (0.3243), reducing the corresponding LSTM baseline (0.8090 RMSE, 0.6457 MAE) by approximately 49.6\% and 49.8\%, respectively. Similarly, the Transformer and RNN versions achieve RMSE reductions of 57.4\% and 38.2\%, respectively, compared to their respective raw-feature baselines.

Each of the three forecasting targets—\textit{Reservation Days}, \textit{Revenue}, and \textit{Number of Reservations}—shows consistent improvement. For example, in the case of Revenue RMSE, the Transformer-based version of our model reduces the error from 1.1040 (baseline) to 0.3502, representing a decrease of over 68\%. Although districts with extremely high Airbnb activity contributed to slightly inflated RMSE values due to large squared errors, our model maintains stable and balanced performance, as evidenced by the consistently lower MAE values.

These improvements validate the effectiveness of our approach. The use of LLM-generated contextual embeddings, combined with the incorporation of external information such as accessibility and human mobility, plays a critical role in enhancing the model's ability to capture complex spatio-temporal patterns. In addition, the integration of label history into the input sequence further stabilizes temporal learning and contributes to substantial accuracy gains.

\subsection{Ablation Study}

To evaluate the contribution of each input modality and the effectiveness of the LLM-based embedding strategy, we conduct two ablation studies on input modalities and LLM-based embedding using the LSTM model, which demonstrates the most consistent performance in our preliminary experiments.

\subsubsection{Ablation on Input Modalities}

Following the architecture shown in Figure~\ref{fig:model}, we test various combinations of input embeddings, including Accessibility, Human Flow, and Airbnb-specific features. All embeddings were generated via LLM encoding as described in Section~\ref{sec:llm_agg}. All configurations include the label embedding for consistency. Table~\ref{tab:ablation_input} reports the corresponding RMSE and MAE scores.

\begin{table}[htbp]
\centering
\caption{Ablation Study on Input Modalities}
\begin{tabular}{lcc}
\toprule
\textbf{Input Data} & \textbf{RMSE} & \textbf{MAE} \\
\midrule
Accessibility                         & 0.5565 & 0.4719 \\
Human Flow                            & 0.4821 & 0.3938 \\
Airbnb                                & 0.7101 & 0.6182 \\
Accessibility + Human Flow           & 0.4226 & 0.3336 \\
Accessibility + Airbnb               & 0.4585 & 0.3675 \\
Human Flow + Airbnb                  & 0.4511 & 0.3635 \\
Accessibility + Human Flow + Airbnb (Our Model) & \textbf{0.4075} & \textbf{0.3243} \\
\bottomrule
\end{tabular}
\label{tab:ablation_input}
\end{table}

Among single-modality inputs, Human Flow features outperform both Accessibility and Airbnb-only configurations. Notably, external variables (e.g., Accessibility and Human Flow) alone achieve better performance than using only Airbnb features, highlighting the importance of contextual information. Performance further improves when combining inputs, with the best results obtained when all modalities are used. These findings demonstrate the effectiveness of semantically enriched embeddings in capturing complex spatio-temporal patterns.

\subsubsection{Effect of LLM-based Embedding}

To assess the impact of LLM-based embeddings, we compare model performance with and without the LLM encoding process. In the ablated version, structured tabular data are used directly, bypassing the prompt construction and LLM embedding stages. Importantly, the input modalities were kept identical in both settings to ensure a fair comparison. Table~\ref{tab:ablation_llm} summarizes the results.

\begin{table}[htbp]
\centering
\caption{Performance Comparison With and Without LLM-based Embedding}
\begin{tabular}{lcc}
\toprule
\textbf{Model Configuration} & \textbf{RMSE} & \textbf{MAE} \\
\midrule
w/o LLM embedding   & 0.6729 & 0.5217 \\
w/ LLM embedding (Our Model) & \textbf{0.4075} & \textbf{0.3243} \\
\bottomrule
\end{tabular}
\label{tab:ablation_llm}
\end{table}

Removing the LLM embedding resulted in significantly worse performance, confirming that the LLM-based embedding approach improves representational capacity by effectively integrating heterogeneous information.

%% file: section/conclusion.tex
\section{Conclusion}

We present a novel regional-level framework for forecasting Airbnb demand, predicting \textit{Reservation Days}, \textit{Revenue}, and \textit{Number of Reservations} using multivariate time-series models. Unlike traditional listing-level approaches, our method integrates regional dynamics and external contextual factors—such as urban accessibility and human mobility—via LLM-based embeddings. Experiments demonstrate that semantically enriched regional representations consistently enhance forecasting performance across RNN, LSTM, and Transformer models, significantly outperforming baseline methods. Our ablation studies further confirm the effectiveness of incorporating contextual features and LLM-based embeddings. This flexible and scalable framework provides practical implications for urban analytics, including zoning policies and tourism demand prediction.

Future research directions include evaluating the generalizability of our framework across diverse urban contexts and investigating its robustness under pandemic-induced anomalies. Moreover, inspired by recent advances in integrating large language models with graph-based reasoning frameworks~\cite{besta2024graph, yao2024chainofthoughteffectivegraphofthoughtreasoning}, we plan to explore GNN+LLM-based architectures to more effectively capture complex spatio-temporal dependencies in regional demand forecasting.